\pdfoutput=1
\documentclass[11pt]{article}

\usepackage{latex/acl}
\usepackage{times}
\usepackage{booktabs} % For prettier tables
\usepackage{latexsym}
\usepackage{graphicx}
\usepackage{adjustbox}
\usepackage{multirow} 
\usepackage{booktabs}

\usepackage{siunitx}
\usepackage{caption}

\usepackage[T1]{fontenc}

\usepackage[utf8]{inputenc}

\usepackage{microtype}

\usepackage{inconsolata}

\title{Optimizing Instruction Synthesis: Effective Exploration of Evolutionary
Space with Tree Search}

\author{
    Chenglin Li$^{1*}$,
    Qianglong Chen$^{1}\thanks{$\quad$ Equal Contribution.}$, 
    \textbf{Zhi Li}$^{1}$,
    \textbf{Feng Tao}$^{1}$,
    \textbf{Yicheng Li}$^{1}$\\
    \textbf{Hao Chen}$^{2}$,
    \textbf{Fei Yu}$^{2}$,
    \textbf{Yin Zhang}$^{1}$\thanks{\quad \ Corresponding author: Yin Zhang.}
    \\
    $^{1}$Zhejiang University, Hangzhou, China \\
    $^{2}$Ant Group, Hangzhou, China \\
    \texttt{\{chenglinli,chenqianglong,zhili,tao\_feng,yichengli,zhangyin98\}@zju.edu.cn } \\
    \texttt{chuhu.ch@antgroup.com, feiyu.fyyu@gmail.com}
}

\begin{document}
\maketitle
% 基于mcts树搜索的指令加工
% Multistep 合成指令 in LLMs with MCTS.
% for Quality, Diversity, and Complexity
% 强调开放域闲聊

\begin{abstract}
Instruction tuning is a crucial technique for aligning language models with humans' actual goals in the real world. Extensive research has highlighted the quality of instruction data is essential for the success of this alignment. However, creating high-quality data manually is labor-intensive and time-consuming, which leads researchers to explore using LLMs to synthesize data. Recent studies have focused on using a stronger LLM to iteratively enhance existing instruction data, showing promising results. Nevertheless, previous work often lacks control over the evolution direction, resulting in high uncertainty in the data synthesis process and low-quality instructions. In this paper, we introduce a general and scalable framework, IDEA-MCTS (Instruction Data Enhancement using Monte Carlo Tree Search), a scalable framework for efficiently synthesizing instructions. With tree search and evaluation models, it can efficiently guide each instruction to evolve into a high-quality form, aiding in instruction fine-tuning. Experimental results show that IDEA-MCTS significantly enhances the seed instruction data, raising the average evaluation scores of quality, diversity, and complexity from 2.19 to 3.81. Furthermore, in open-domain benchmarks, experimental results show that IDEA-MCTS improves the accuracy of real-world instruction-following skills in LLMs by an average of 5\% in low-resource settings.
\end{abstract}
\begin{figure}
    \centering \includegraphics[width=1.0\linewidth]{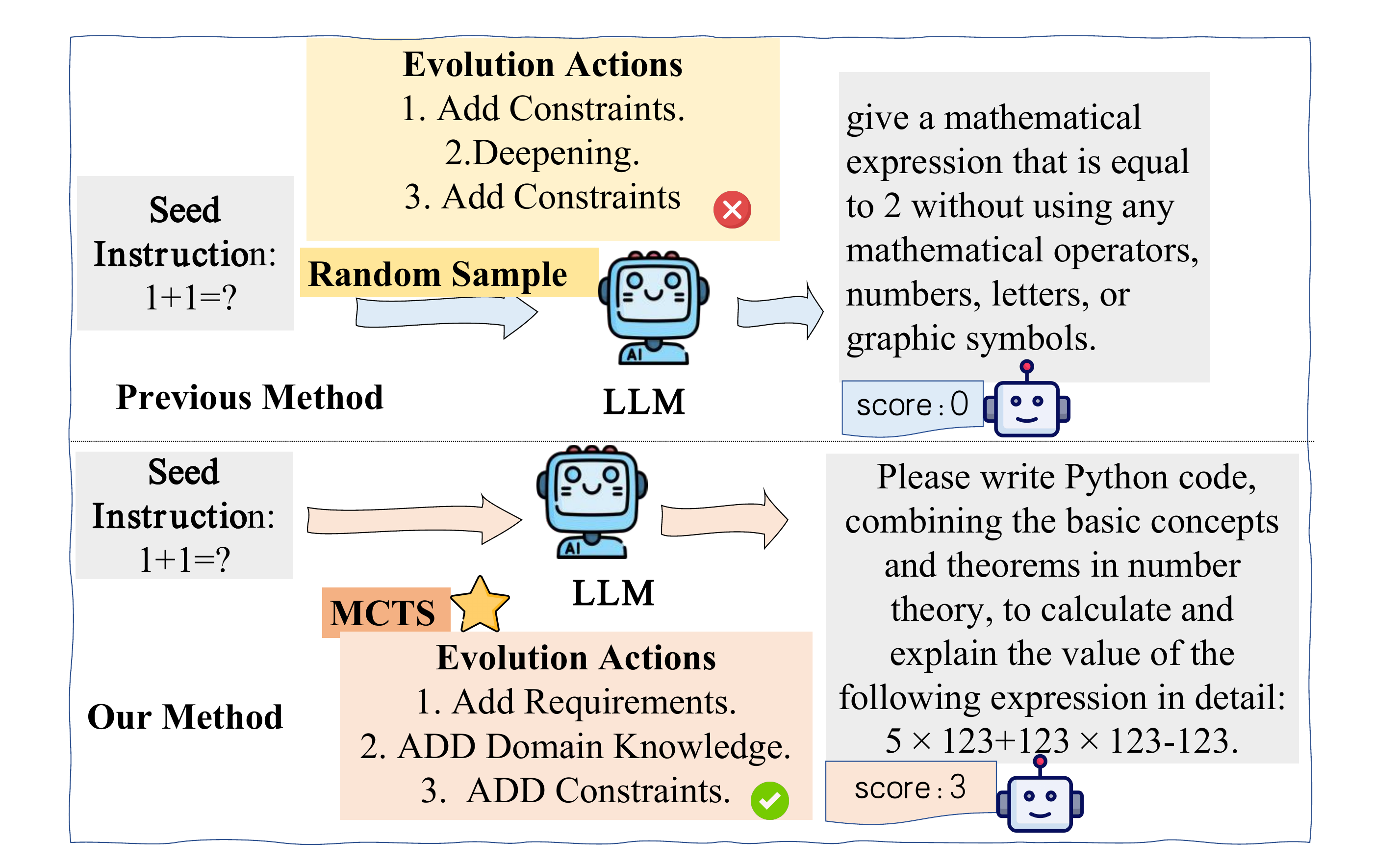}
    \caption{Iteratively enhance seed instructions using LLMs: The prior method's random sampling instruction evolution led to a perplexing instruction by selecting ``Add constraints''  multiple times. Our method uses MCTS to find suitable prompts, resulting in high-value instructions that align the language model to effectively learn multiple skills.}
    \label{fig:methods}
\end{figure}
\begin{figure*}
    \centering \includegraphics[width=1.0\linewidth]{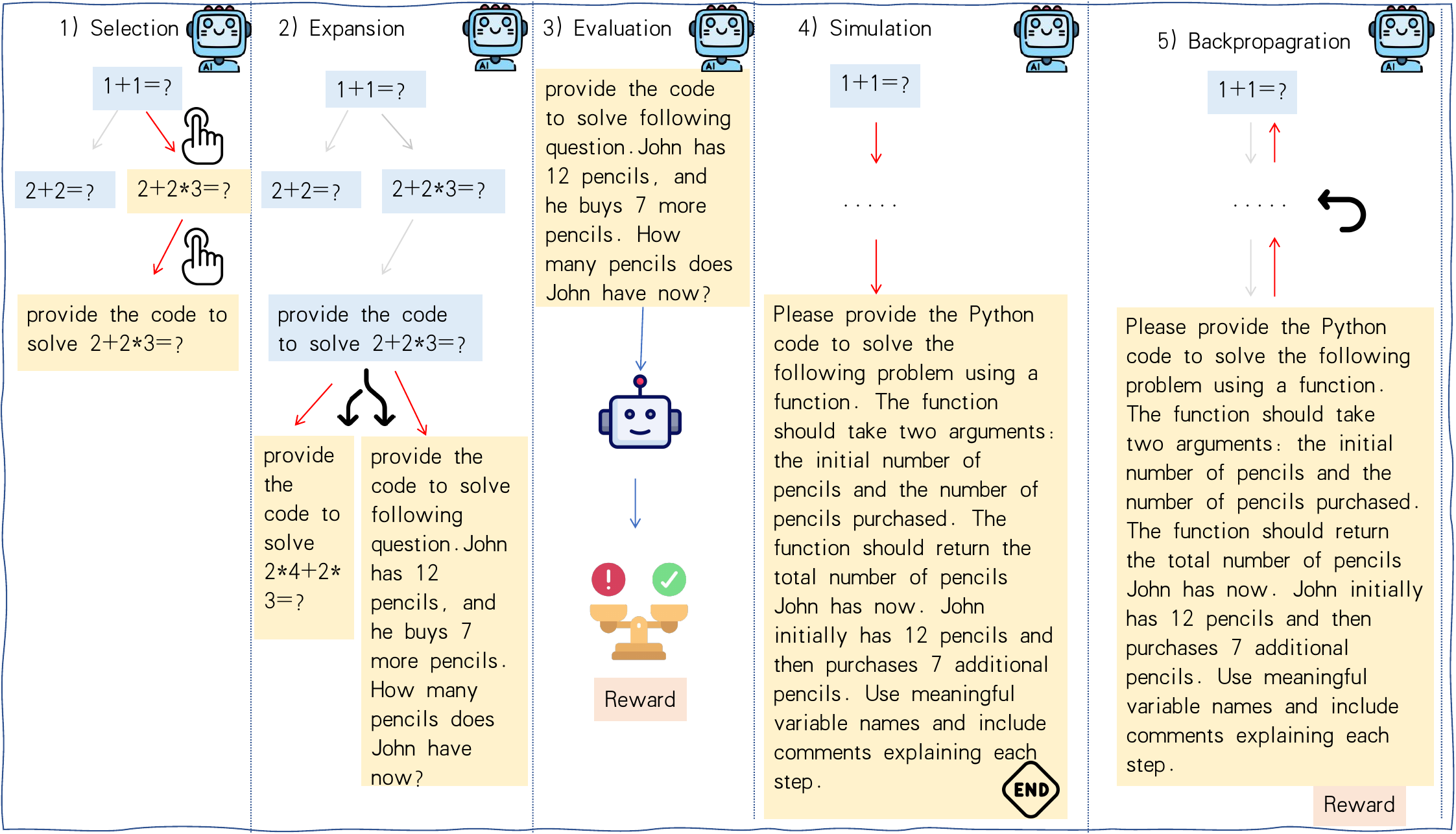}
    
    \caption{Framework of MCTS for instruction synthesis: 1. \textbf{Selection}: Choose high-value leaf nodes. 
2. \textbf{Expansion}: Evolve the selected leaf nodes to generate new nodes. 
3. \textbf{Evaluation}: Assess the current node to determine a reward. 
4. \textbf{Simulation}: Randomly evolve the current instruction to a terminal state. 
5. \textbf{Backpropagation}: Propagate the terminal state's reward back through the path's nodes.}
    \label{fig:framework}
\end{figure*}

% Progress in this area highlights the critical role of data quality, complexity, and diversity in enhancing the instruction-following capabilities of LLMs~\cite{marion2023less,li2023quantity,cao2023instruction,chen2023maybe,wu2023self,bukharin2023data,zhao2023preliminary}.

\section{Introduction}
Large language models (LLMs) have exhibited remarkable capabilities across a wide range of tasks in the field of natural language processing (NLP)~\cite{brown2020language,kojima2022large,wei2022chain,ouyang2022training,touvron2023LLaMA,jiang2023mistral, OpenAI2023GPT4TR}. Notably, LLMs can be trained to enhance their instruction-following skills through various methods, including fine-tuning on human-annotated data~\cite{ouyang2022training,zhou2023lima,touvron2023LLaMA} or extracted knowledge from stronger LLMs~\cite{wang2022self,xu2023wizardlm,zhao2023preliminary,xu2023wizardlm,xu2023rethinking,wang2024survey}.~\citet{zhou2023lima} have demonstrated that this alignment can be achieved with low-resource 1k data. However, acquiring such data through human annotation remains high-cost, thus limiting further progress.

Recent work explores synthesizing instruction data with LLMs by prompting them with example data or prompts and iteratively enhancing the instruction data, offering an efficient and cost-effective alternative to human annotation~\cite{xu2023wizardlm,luo2023wizardcoder,luo2023wizardmath,liu2023makes}. They introduced evolution prompts for LLMs, such as ``Add constraints'', ``Increase reasoning'' and ``Complete input.'', enabling LLMs to iteratively improve seed instructions. However, the process suffers high uncertainty due to the limited evolution prompts, random selection methods, and lack of control over the evolution direction. Specifically, failures occur when LLMs select inappropriate evolution prompts or fail to halt the instruction synthesis process appropriately. As shown in Figure \ref{fig:methods}, randomly selecting the prompts can turn a seed instruction like "1+1=" into a perplexing instruction. Language models will struggle to learn new skills from these low-value instructions, as humans also find them difficult to understand. Conversely, a few high-value instructions can significantly enhance the model's skills, enabling it to solve real-world problems.

Intuitively, simple seed instructions can evolve into a wide variety of forms during the evolutionary process. To efficiently optimize and control this evolution, we introduce a novel framework, IDEA-MCTS, which expands the evolution prompts as the action space and incorporates a tree search algorithm to iteratively enhance seed instruction data. In MCTS, each seed instruction acts as the root node. High-value nodes are identified through selection and use evolution prompts for further expansion, followed by simulation and backtracking, to find an optimal evolution action space to enhance the instructions. In this process, we employ customizable evaluation models to assess the quality, diversity, and complexity of the nodes, effectively controlling the direction of instruction evolution. This framework enhances instruction data and provides a clearer understanding of the evolution process, as shown in the case analysis in Appendix \ref{case_study}. Our experimental results show that IDEA-MCTS significantly enhances the seed instruction data and models fine-tuned on this enhanced data exhibit substantial improvements compared to previous methods. We believe this work provides clear guidance for instruction synthesis, aiding models in achieving data-efficient alignment and enhancing overall performance. The contributions of our work are as follows:
\begin{itemize}
% 之前mcts主要
\item To synthesize high-value instructions for enhancing language model skills, we propose IDEA-MCTS, a scalable framework that controls the direction of instruction evolution by expanding the evolution space and integrating evaluation models in tree search.

% evlotion space,直观的提供了一种扩展进化动作空间的方法，并通过实验进行了验证其有效性
\item To enhance the efficiency and accuracy of instruction evolution, we expand the existing limited evolutionary space in two ways: evolving general effective instructions from themselves, and evolving task-specific instructions by designing meta-prompts. 
 
\item We demonstrate the effectiveness of our framework by analyzing the generated instructions and fine-tuning open-source models, including LLaMA2, LLaMA3, Phi-3, and Mistral, across different seed datasets and tasks, achieving a 5\% improvement over the previous random evolution method on the open-domain instruction-following benchmark.
\end{itemize}

\section{Related Work}
\paragraph{Data Synthesis for Instruction Tuning}
Instruction tuning (IT) is a crucial technique for enhancing the performance and alignment of LLMs~\cite{taori2023stanford,chiang2023vicuna,wang2023aligning}. Recent efforts have extended into open-domain IT, characterized by a wide range of formats and task types, driven by crowdsourced human-generated instruction-response pairs~\cite{kopf2304openassistant,conover2023free,zhang2023instruction,peng2023instruction,zhou2023lima}. However, the high cost of human annotation poses significant challenges~\cite{zhang2023instruction}.  One promising solution for this limitation is the synthesis of instruction data with the help of stronger LLMs~\cite{bai2022constitutional,OpenAI2023GPT4TR,Anil2023PaLM2T,geminiteam2023gemini}. Yet, using LLM-generated data increases the risk of low-quality examples, highlighting the need for more focus on dataset refinement and enhancement.
Some works~\cite{chen2023alpagasus,lu2023instag,liu2023makes} address this by prompting stronger LLMs to filter instruction data based on its quality, diversity, and complexity, serving as a form of refinement. However, this approach lacks the synthesis of new instruction, limiting the model's instruction-following capabilities, especially in low-resource scenarios where only a small amount of data is available. 
Other works~\cite{zhao2024tree,xu2023wizardlm} enhance existing seed instructions by using LLMs with carefully designed prompt templates.~\citet{zhao2024tree} enhanced the original instructions using tree-structured prompts but focused only on the complexity and heavily relies on LLMs' intrinsic knowledge. Additionally, some work~\cite{xu2023wizardlm,luo2023wizardcoder,luo2023wizardmath,liu2023makes} design a series of evolution prompts to iteratively guide LLMs in enhancing the seed instructions. However, random selection during instruction evolution introduces high uncertainty and affects the quality of generated instructions. To effectively enhance the seed instruction data, we propose IDEA-MCTS, which expands the evolution action space, introduces evaluation models and iteratively improves instruction data with MCTS.

\paragraph{Tree Search for LLM Enhancement}
Tree search methods such as BFS, A* search~\cite{hart1968formal}, and MCTS~\cite{kocsis2006bandit,coulom2006efficient,ye2021mastering,silver2016mastering}, are widely used to find an optimal state in a tree structure. Integrating tree-search methods with LLMs presents a novel approach to find an effective sequence of actions that leads to a favorable outcome. Effective search strategy is crucial for reasoning and planning~\cite{hao2023reasoning,zhou2023language,hu2023tree}. Depth/breadth-first search in ~\cite{yao2023tree}, A* search in~\cite{zhuang2023toolchain} and MCTS in~\cite{zhang2023planning,yu2023prompt,hao2023reasoning,zhou2023language,chen2024tree}. ~\citet{feng2023alphazero,tian2024toward,chen2024alphamath} have utilized tree search for LLM self-improvement. Unlike previous approaches, we leverage the powerful generative capabilities of LLMs and MCTS for instruction synthesis.
\begin{figure*}
    \centering \includegraphics[width=1.0\linewidth]{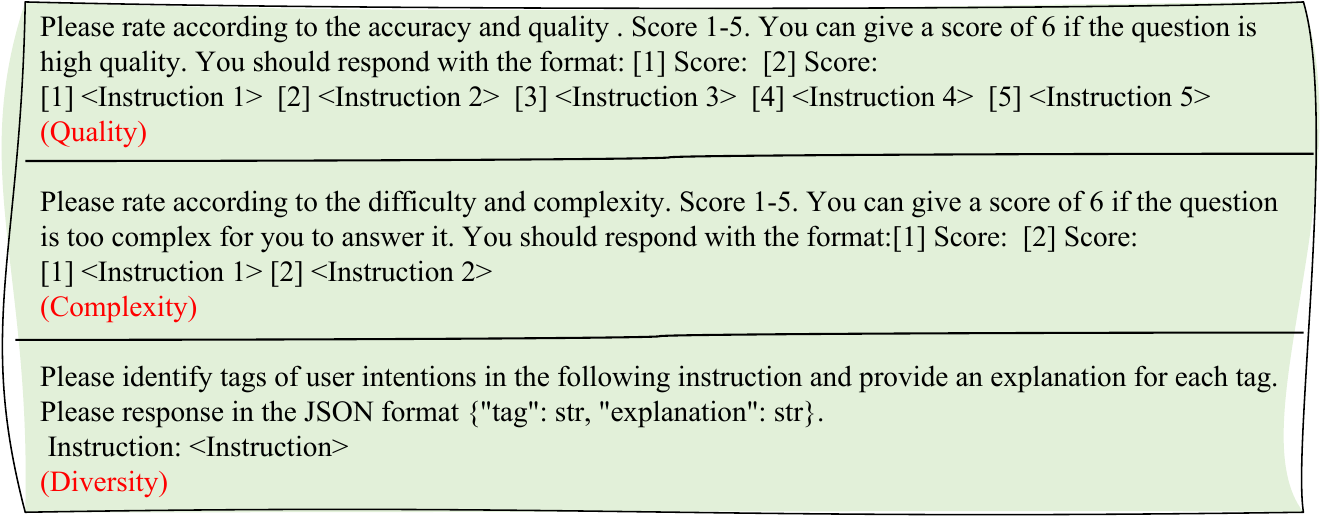}
    \caption{Evaluation prompt used to assess the quality, complexity, and diversity of instructions. Instruction diversity is measured by the number of distinct intents.}\label{figure:eva_prompt}
\end{figure*}

\begin{table*}[ht]
    \centering
    \renewcommand{\arraystretch}{1.2} % 调整行高
    \adjustbox{width=\textwidth}{
        \begin{tabular}{|>{\centering\arraybackslash}m{0.3\textwidth}|>{\arraybackslash}m{0.65\textwidth}|}
            \hline
            \textbf{Evolution Space} & \textbf{Description} \\ \hline
            Add Global and Local Goals & Add one or more global and local goals into the instruction to enhance its direction and purpose. \\ \hline
            Add Key Constraints & Add one or more constraints where necessary to define its limitations and boundaries. \\ \hline
            Add Task Requirements & Specify one or more detailed requirements to clarify the tasks within the instruction. \\ \hline
            Add Problem-Solving Skills & Add one or more problem-solving task skills. \\ \hline
            Add Reasoning Complexity & Increase complexity by adding one or more reasoning elements. \\ \hline
            Add Domain Knowledge & Add one or more areas of domain-specific knowledge, such as medicine, law, finance, IT technology. \\ \hline
            Add Life Topics & Add one or more life topics. Topics can range from health and nutrition, cooking, photography, music, and travel, to parenting. \\ \hline
            Add Real-World Applications & Add one or more real-world applications to provide practical context and applicability, such as education, customer service, and Business. \\ \hline
            Add Emotional Expression & Add one or more emotional content elements such as excitement or concern. \\ \hline
            Format the Input Style & Define one or more input formatting styles, such as a doctor, teacher, or customer. \\ \hline
            Format the Output Style & Specify one or more output formats, such as report format or summarized in paragraphs. \\ \hline
            Create a New One & Create one instruction within the same domain to introduce fresh perspectives. \\ \hline
        \end{tabular}
    }
    \caption{Expanded Space for Instruction Evolution.}
    \label{table:attr and action}
\end{table*}
\section{Approach} 
In this section, we introduce the novel framework IDEA-MCTS, which enhances the quality, diversity, and complexity of seed instructions with a stronger LLM, using MCTS. We first define the problem, including the state, action space, and reward function. Then, we discuss the expansion of evolution prompts from two key aspects and the use of MCTS with LLM to efficiently explore the action spaces. Finally, we fine-tune models based on the instruction data generated by the LLM, proving the effectiveness of the framework in low-resource settings.

\subsection{Problem Setting}
We begin with a seed instruction sample $x$ as the root node and employ a stronger language model $p_{\theta}$. Our goal is to improve the quality, diversity, and complexity of $x$. To achieve this, we use evolution prompts, such as `add constraints', as our action space. During the tree search, intermediate instructions generated by the LLM, denoted as $z_t$, serve as new nodes. 
\begin{equation}
z_{t+1} = p_{\theta}(z_t, a)
\end{equation}
By applying an action $a$, which is an evolution prompt to wrap the state $z_t$, we obtain the next instruction $z_{t+1}$ via $p_{\theta}$.
We assess each intermediate instruction $z_t$ based on its quality, diversity, and complexity. The value $v(z_t)$ of an instruction is determined using the following equation:
\begin{equation}
v(z_t) = p_{\theta_q}(z_t) + p_{\theta_d}(z_t) + p_{\theta_c}(z_t)
\end{equation}

In this equation, $p_{\theta_q}(z_t)$, $p_{\theta_d}(z_t)$, and $p_{\theta_c}(z_t)$ represent the quality, diversity, and complexity scores of the instruction $z_i$, respectively. Notably, instruction diversity is measured by the number of distinct intents. Further details about these value scores will be discussed in the following sections. By integrating these elements, we aim to create a framework that robustly enhances seed instructions.

\paragraph{Quality \& Complexity \& Diversity}

Following the \cite{liu2023makes,lu2023instag}, we continue training based on models, EVOL\_QUALITY, EVOL\_COMPLEXITY, and InsTagger from these works with 1k data points. We apply a random evolution method\cite{xu2023wizardlm} to create new data points from a base sample, gradually adjusting their complexity, quality, and diversity of instruction. We evaluate these data points using ChatGPT and train an automatic scoring model with LLaMA2-7B to predict ChatGPT's scores. The evaluation prompt we use is shown in Figure \ref{figure:eva_prompt}. These scoring models are used to assess the quality, complexity, and diversity of instructions as rewards in MCTS. 

 \subsection{Instruction Evolution with MCTS}

In our framework, we leverage a stronger language model \( p_\theta \) and value function \( v \) to evolve the seed instruction \( x \) using MCTS, as shown in Figure \ref{fig:framework}. 

\begin{figure}
    \centering \includegraphics[width=0.85\linewidth]{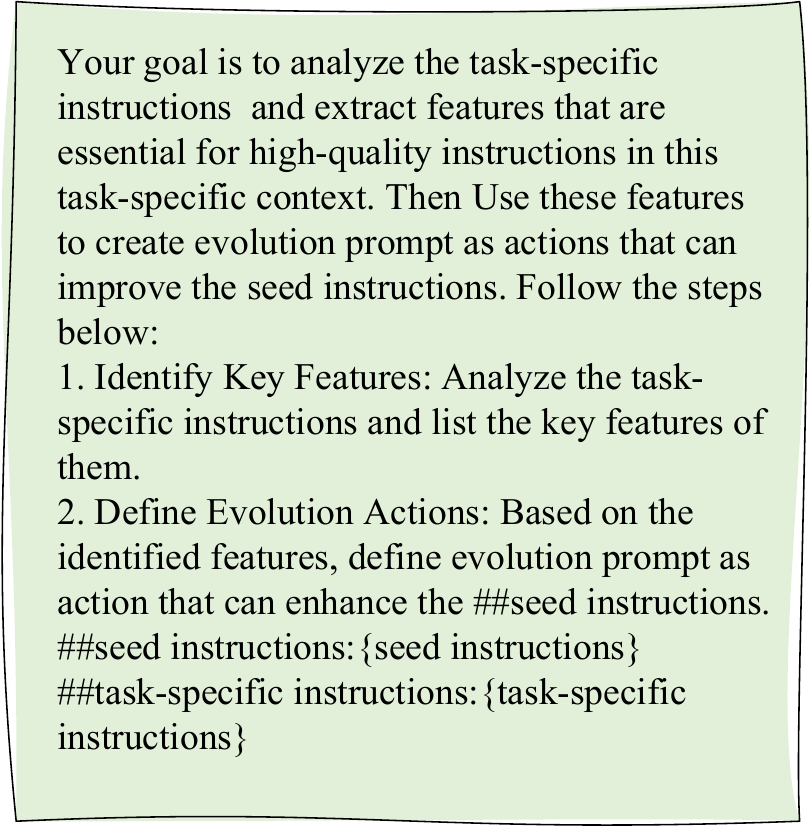}
    \caption{Meta prompt used in LLM: Extracting evolutionary prompt actions from task-related contexts.}
    \label{fig:meta_prompt}
\end{figure}

Intuitively, more precise and diverse evolution prompts contribute to enhancing the quality of seed instructions. To achieve this, we first expand the evolution prompts from two ways, general effective and task-specific instructions. We explore the open-space evolution prompts, that contribute a general effective instructions such as goals, key constraints, and requirements \cite{xu2023wizardlm}. On the other hand, we aim to ensure that the seed instructions can effectively transfer to task-specific contexts. With LLMs, we design the meta prompts, as shown in Figure \ref{fig:meta_prompt}, to extract task-related evolution prompts that contain the words "such as." 
As shown in Table \ref{table:attr and action}, the designed evolution prompts can enhance both the depth and breadth of the seed instruction. 

% We show more details about the evolution prompts in Appendix \ref{evol_prompt}.

Then we construct a decision tree. MCTS proceeds for \( k \) episodes, starting from the root (initial state) and progressively expanding this tree through two primary steps: Selection and Expansion. During \textbf{Selection}, the child with the highest Upper Confidence bounds applied to Trees (UCT) value~\cite{kocsis2006bandit,coulom2006efficient} is chosen for the next iteration. The UCT of a child state \( z \) is computed as follows:
\begin{equation}
UCT(z) = V(z) + C \cdot \sqrt{\frac{\ln(N(p))}{N(z)}}
\end{equation}
where \( N(z) \) represents the number of visits to node \( z \), and \( V(z) \) is the value function (expected return). During \textbf{Expansion}, multiple child states \( z \) are explored from the current state \( p \) by sampling \( n \) actions. The child node with the highest UCT value is selected for expansion in the subsequent iteration. In \textbf{Evaluation}, we assess the quality, complexity, and diversity of the instruction data using the value function \( v \), which serves as the node's reward. In \textbf{Simulation}, selection and expansion are performed repeatedly until a termination state is reached, constructing the rollout policy. The termination state occurs when the tree's depth or node value meets a specified threshold. \textbf{Backpropagation} is performed at the end of an episode: the return \( v \) is used to update every \( V(z) \) along the path using the formula:
\begin{equation}
V(z) = V_{\text{old}}(z) \cdot \left( \frac{N(z) - 1}{N(z)} \right) + \frac{v}{N(z)}
\end{equation}
where \( V_{\text{old}}(z) \) denotes the old value function.

MCTS relies on an environment model to reverse steps and build a search tree, imposing strict assumptions. This constraint does not apply to LLMs. Our method allows resetting to any step by copying historical text input, overcoming the limitation. By integrating MCTS with LLMs, we demonstrate how heuristic search algorithms can efficiently evolve instructions by leveraging the powerful generative capabilities of LLMs. 

Finally, after evolving the seed instructions, we obtain responses from the stronger LLM and fine-tune the open-source model. To ensure clarity and logic, we avoided complex templates from previous works~\cite{wei2021finetuned,longpre2023flan}. Instead, our method follows a straightforward instruction template~\cite{taori2023stanford}.

\section{Experiments}
% In this section, we first introduce the experiment settings and then analyze the enhanced instruction data. Finally, we present model performance after training with enhanced instructions. Additionally, we conduct a case study
% in Appendix \ref{case_study} to show how IDEA-MCTS enhances the seed instruction.

\subsection{Experiments Setting}
\paragraph{Baselines}
We compare our method with manually annotated data and other techniques for enhancing instructional data using a more powerful LLM. Additionally, we present the baselines utilized in our experiments:

\begin{itemize}
    \item \textbf{Seed}: Serves as the baseline without any enhancement methods.
    
    \item \textbf{LIMA}~\cite{jha2023limit}: Contains 1,000 high-quality, human-annotated instructional data points, demonstrating significant improvements for LLMs.
    
    \item \textbf{Tree-instruct}~\cite{zhao2024tree}: Enhances the complexity of instructional data by adding nodes to the semantic tree structure.
    
    \item \textbf{Evol-Instruct}~\cite{xu2023wizardlm,luo2023wizardmath,luo2023wizardcoder}: Distinguishes itself by prompting the LLM to iteratively evolve instructional data in a random step-by-step manner.

 \end{itemize}

\paragraph{Test Datasets}
Many studies have focused on assessing the capabilities of LLMs~\cite{liang2022holistic}. However, the challenge remains unresolved. A prevalent method involves using the powerful language model as the evaluator~\cite{li2023alpacaeval,zheng2024judging,chiang2023vicuna,chen2023alpagasus}. 
In our framework, we employ two distinct methods to assess the model's capabilities: LLM evaluation and human evaluation. Specifically, we use Alpaca-Eval 
\cite{alpaca_eval} and MT-Bench \cite{zheng2024judging} to assess real-world instruction-following capabilities. We show more details in Appendix \ref{Appendix_benchmark}. In the Alpaca-Eval, we compare our model's output with Text-Davinci-003 and use GPT-3.5-turbo to evaluate and score the output in MT-Bench. Additionally, we evaluate the model's capabilities in the NLP benchmark with the OpenLLM Leaderboard, which comprises four tasks: ARC~\cite{clark2018think}, HellaSwag~\cite{zellers2019hellaswag}, GSM8K~\cite{cobbe2021training}, and TruthfulQA~\cite{lin2022truthfulqa}. 

\begin{table}[]
\adjustbox{width=0.48\textwidth}{
\begin{tabular}{ccccc}
\toprule
\multirow{2}{*}{Method} & \multicolumn{4}{c}{Metric} \\ \cmidrule(lr){2-5}
 & Quality & Instag & Complexity & Average \\ 
\midrule
Seed & 3.58 & 1.60 & 1.40 & 2.19 \\
Lima & 3.58 & 1.99 & 2.09 & 2.55 \\
Tree-instruct & 4.37 & 2.40 & 2.44 & 3.07 \\
Evol-Instruct & 3.82 & 2.30 & 2.52 & 2.87 \\
Evol-Instruct+ & 4.01 & 2.80 & 2.70 & 3.17 \\
MCTS & 3.96 & 3.12 & 3.51 & 3.53 \\
MCTS+ & \textbf{4.56} & \textbf{3.24} & \textbf{3.62} & \textbf{3.81} \\
\bottomrule
\end{tabular}
}

\caption{Statistics of instruction dataset. The "+" symbol indicates methods that expand the evolution prompts space.}
\label{Table:statistics}

\end{table}

\begin{table*}
    \centering
    % \vspace{-40pt}
    \adjustbox{width=0.75\textwidth}{
    \begin{tabular}{llcccccc}
        \toprule
        \multirow{2}{*}{Model} & \multirow{2}{*}{Method} & \multicolumn{5}{c}{Metrics} \\
        \cmidrule(lr){3-7}
        & & help\_base & koala & self\_instruct & oasst & vicuna & overall \\
        \midrule
        \multirow{9}{*}{LLaMA2} & Seed & 39.53 & 49.36 & 38.10 & 55.85 & 45.00 & 45.59 \\
        & Lima & 44.96 & 42.95 & 30.95 & 51.60 & 33.75 & 40.68 \\
        % & Mixed\_LIMA\_Alpaca & 42.64 & 44.87 & 39.29 & 52.13 & 36.25 & 43.66 \\
        %     & Alpaca\_52k\_Top\_1k & 41.09 & 48.72 & 41.76 & 56.91 & 40.00 & 46.34 \\
  
         & Evol-Instruct & 44.96 & 45.51 & 43.65 & 51.60 & 42.50 & 46.46 \\
          % & Evol-Instruct\_Top\_1k & 45.74 & 52.56 & 46.83 & 60.11 & 38.75 & 50.06 \\
      
        & MCTS+ (ours) & 51.94 & 50.64 & 45.24 & 63.83 & 42.50 & \textbf{51.61} \\
        \midrule
        \multirow{5}{*}{LLaMA3} & Seed & 58.14 & 54.49 & 51.98 & 64.89 & 47.50 & 56.27 \\
        & Lima & 44.19 & 41.67 & 38.10 & 52.13 & 36.25 & 42.98 \\
        & Tree-instruct & 53.49 & 57.05 & 55.95 & 67.55 & 52.50 & 58.21 \\
        & Evol-Instruct & 56.59 & 52.56 & 51.98 & 68.09 & 52.50 & 57.08 \\
        & MCTS+ (ours) & 53.49 & 61.54 & 62.70 & 70.21 & 46.25 & \textbf{60.37} \\
        \midrule
        \multirow{5}{*}{Phi-3} & Seed & 46.51 & 55.13 & 55.16 & 64.89 & 43.75 & 55.22 \\
        & Lima & 41.86 & 50.64 & 41.27 & 60.11 & 43.75 & 48.36 \\
        & Tree-instruct & 51.94 & 57.05 & 51.59 & 69.08 & 51.25 & 57.52 \\
        & Evol-Instruct & 51.94 & 55.77 & 51.19 & 68.09 & 46.25 & 56.09 \\
        & MCTS+ (ours) & 57.36 & 57.05 & 59.52 & 72.34 & 65.00 & \textbf{62.36} \\
        \midrule
        \multirow{5}{*}{Mistral} & Seed & 56.59 & 55.77 & 52.78 & 70.21 & 45.00 & 57.52 \\
        & Lima & 41.86 & 48.08 & 42.86 & 53.72 & 36.25 & 45.59 \\
        & Tree-instruct & 55.04 & 59.62 & 53.57 & 69.62 & 46.25 & 58.39 \\
        & Evol-Instruct & 57.36 & 55.77 & 52.78 & 70.21 & 41.25 & 57.52 \\
        & MCTS+ (ours) & 61.24 & 60.90 & 58.33 & 72.87 & 58.75 & \textbf{62.80} \\
        \bottomrule
    \end{tabular}
}
    \caption{Results of different instruction-tuned models on Alpaca-Eval (\%).}
    \label{Alpaca-eval_main}
    % \vspace{-15pt}
\end{table*}

\paragraph{Experiment Setting}
We randomly select 1,000 seed instructions each from Alpaca-52K \cite{taori2023stanford} and Dolly \cite{conover2023free} as a low-resource setting. We initialize our MCTS evolution process with the stronger LLM, GPT-3.5-turbo-0125 model. When calling this API, we define the temperature parameter to 0.7, set the hyperparameter C to 1 set the maximum token limit to 2048, and apply no penalty. In the MCTS setup, we generate evolution prompts for the seed instructions based on the task-specific Alpaca-eval benchmark. The terminal state is defined as either reaching a depth of more than 4 or achieving a reward of more than 10. For each iteration, we expand 5 nodes per step, and the MCTS process is iterated 3 times. We randomly select 1,000 data points from the generated data, collected from paths between the root node and terminal state nodes, as training data for the low-resource setting. During tuning, the foundational models for our experiments are the LLaMA2-7B, Mistral-7B, LLaMA3-8B, and Phi-3. To efficiently fine-tune these models, we adopted the QLORA approach \cite{dettmers2023qlora}. Throughout the tuning process, we maintained a batch size of 32 and ended the process after a maximum of 800 training steps. It's important to note that these preliminary experiments were conducted on a single GPU with 48GB of memory. For technical execution, we harnessed the capabilities of HuggingFace Transformers, PyTorch, and Accelerate, ensuring strict adherence to academic integrity and standards throughout the entire process.

\subsection{Statistical Analysis of the Data Evolved from MCTS}
\label{ss:statics}
We conduct a comprehensive analysis of the evolved instruction data from three critical dimensions: quality, complexity, and diversity with the EVOL\_QUALITY, EVOL\_COMPLEXITY, and InsTagger. As shown in Table \ref{Table:statistics}, the Seed contains 1,000 instructions selected from the Alpaca-52K. The Evol-Instruct contains 1,000 instructions obtained through random evolution, while the MCTS contains 1,000 instructions obtained through the MCTS evolution. MCTS+ method can achieve the highest scores across all evaluation metrics~\cite{liu2023makes,lu2023instag}, demonstrating significant improvement in quality, diversity, and complexity. It outperforms the Seed, with average scores increasing from 2.19 to 3.81. The expansion of the instruction evolution space proves to be a highly effective strategy for enhancing the quality of instruction data.

\begin{table}
    \centering
    \adjustbox{width=0.45\textwidth}{
    \begin{tabular}{lccc}
        \toprule
        & \multicolumn{3}{c}{MT-Bench} \\
        \cmidrule(r){2-4}
        & Turn-1 & Turn-2 & Average Score \\
        \midrule
        Seed    & 6.25 & 5.54 & 5.90 \\
        Lima    & 6.71 & 6.61 & 6.66 \\
         Tree-instruct    & 6.55 & 6.03 & 6.29 \\
          Evol-Instruct  & 6.63 & 6.45 & 6.54 \\
           MCTS    & 6.71 & 6.61 & 6.66 \\
            Evol-Instruct+ & 6.56 & 7.14 & 6.69 \\
             MCTS+  & 6.74 & 7.14 & \textbf{6.94} \\
        \bottomrule
    \end{tabular}
    }
    \caption{Results of different instruction-tuned models on MT-Bench.}
    \label{tab:mt_bench_results}
\end{table}

\begin{table*}
    \centering
    \adjustbox{width=0.85\textwidth}{
    \begin{tabular}{llcccccc}
        \toprule
        \multirow{2}{*}{Model} & \multirow{2}{*}{Method} & \multicolumn{5}{c}{Metrics} \\
        \cmidrule(lr){3-7}
        & & help\_base & koala & self\_instruct & oasst & vicuna & overall \\
        \midrule
        \multirow{5}{*}{LLaMA2} 
        % & Seed & 28.68 & 24.36 & 37.77 & 20.24 & 22.50 & 26.89 \\
        & Tree-instruct & 51.16 & 51.28 & 60.11 & 42.46 & 43.75 & 49.81 \\
        & Evol-Instruct & 44.19 & 49.36 & 60.11 & 42.46 & 41.25 & 48.07 \\
        & MCTS & 44.96 & 51.92 & 62.23 & 44.84 & 41.25 & 50.00 \\
        & Evol-Instruct+ & 50.39 & 53.21 & 59.57 & 42.86 & 37.50 & 49.50 \\
        & MCTS+ & 49.61 & 51.92 & 63.30 & 44.84 & 42.50 & \textbf{51.18} \\
        \bottomrule
    \end{tabular}
}
    \caption{Results of different instruction-tuned models on Alpaca-Eval using Dolly as the Seed Dataset (\%).}
    \label{Alpaca-eval:dolly}
\end{table*}

\begin{table*}[h]
\centering
 \adjustbox{width=0.85\textwidth}{
\begin{tabular}{llcccccc}
    \toprule
    Model & Method & ARC-Easy & ARC-Challenge & HellaSwag & TruthfulQA & GSM8k & Average \\
    \midrule
    \multirow{5}{*}{LLaMA2} 
     & Seed & 80.30 & 52.82 & 77.96 & 29.80 & 13.04 & 50.78 \\
                            & Lima & 80.43 & 53.07 & 78.54 & 31.18 & 14.10 & 51.59 \\
                            & Tree-instruct & 80.85 & 53.92 & 78.80 & 33.09 & 13.80 & 52.01 \\
                            & Evol-Instruct & 80.81 & 54.10 & 78.81 & 32.51 & 13.50 & 51.95 \\
                            % & MCTS & 79.88 & 54.10 & 78.68 & 33.60 & 13.72 & 52.00 \\
                            % & WizardLM+ & 81.23 & 53.92 & 78.89 & 33.09 & 13.42 & 52.11 \\
                             & MCTS+ & 81.02 & 54.10 & 78.94 & 33.73 & 13.72 & \textbf{52.30} \\
                      & MCTS+ & 84.97 & 62.43 & 78.54 & 42.14 & 74.60 & 68.52 \\
    \bottomrule
\end{tabular}
}
\caption{Results of different instruction-tuned models on the NLP benchmark, OpenLLM (\%).}
\label{table:openllm}
\end{table*}

\subsection{Main results}
\label{ss:results}
The main results presented below are based on LLM evaluations and further human evaluations are provided in Appendix \ref{human_eval}.

Table \ref{Alpaca-eval_main} demonstrates that models fine-tuned with data evolved from MCTS+ exhibit better performance compared to other fine-tuning methods. In particular, LLaMA2 and LLaMA3 can show significant gains with MCTS+, with improvements of 6.02\% and 4.1\%, respectively, over the Seed method. Furthermore, Phi-3 and Mistral fine-tuned with  MCTS+ method outperform previous methods across various skills, including help\_base, koala, self\_instruct, oasst, and vicuna. Notably, the Mistral model achieves a win rate of 61.24\% in help\_base, surpassing the previous highest win rate by 3.88 obtained using the Evol-Instruct method. Overall, Mistral exhibits a 5.28\% enhancement in performance compared to the Evol-Instruct method. These results show that MCTS effectively enhances models' instruction-following capabilities better than traditional methods. Additionally, fine-tuning with the LIMA method does not significantly improve the model's performance on Alpaca-eval, suggesting potential generalization limitations of manually annotated models. 

\subsection{Generalization}
During the expanded evolution process, with a focus on task-specific instruction data features on Alpaca-Eval, we also evaluate the model's performance on the open-domain benchmark MT-Bench and assess its capabilities on the NLP benchmark, OpenLLM. Additionally, we consider the effectiveness of using Dolly as a seed dataset.

As shown in Table \ref{tab:mt_bench_results}, the MCTS+ method enhances both the model's single-turn and multi-turn dialogue capabilities. The single-turn score is improved from 6.25 (Seed) to 6.74 (MCTS+), while the multi-turn score is increased from 4.54 (Seed) to 7.15 (MCTS+). This results in an overall average score improvement from 5.90 (Seed) to 6.94 (MCTS+), highlighting the method's effectiveness in handling more complex, multi-turn dialogues.

Using Dolly as the seed instruction dataset, Table \ref{Alpaca-eval:dolly} shows that the MCTS+ method can achieve the best performance, with a 3\% improvement compared to the Evol-Instruct method. Specifically, the overall score is improved from 48.07\% (Evol-Instruct) to 51.18\% (MCTS+). In individual metrics, MCTS+ can improve the help\_base from 44.19 to 49.61, koala from 49.36\% to 51.92\%, self\_instruct from 60.11\% to 63.30\%, oasst from 42.46\% to 44.84\%, and vicuna from 41.25\% to 42.50\%. 

As shown in Table \ref{table:openllm},
despite being fine-tuned on very different instruction-following prompts, the model's capabilities in NLP tasks show a slight improvement, with a 1.5\% increase compared to the seed method.

\begin{table}
    \centering
    \adjustbox{width=0.3\textwidth}{
    \begin{tabular}{llc}
        \toprule
        Model & Method & Overall (\%) \\
        \midrule
        \multirow{4}{*}{LLaMA2} 
        & Evol-Instruct & 46.46 \\
        & MCTS & 47.89 \\
        & Evol-Instruct+ & 49.32 \\
        & MCTS+ & \textbf{51.61} \\
        \midrule
        \multirow{4}{*}{LLaMA3} 
        & Evol-Instruct & 57.08 \\
        & MCTS & 57.52 \\
        & Evol-Instruct+ & 58.12 \\
        & MCTS+ & \textbf{60.37} \\
        \midrule
        \multirow{4}{*}{Phi-3} 
        & Evol-Instruct & 56.09 \\
        & MCTS & 57.64 \\
        & Evol-Instruct+ & 59.44 \\
        & MCTS+ & \textbf{62.36} \\
        \midrule
        \multirow{4}{*}{Mistral} 
        & Evol-Instruct & 57.52 \\
        & MCTS & 57.57 \\
        & Evol-Instruct+ & 60.11 \\
        & MCTS+ & \textbf{62.80} \\
        \bottomrule
    \end{tabular}
    }
    \caption{Ablation study results on Alpaca-Eval (\%).}
    \label{Ablation-study}
    % \vspace{-15pt}
\end{table}

% 针对随机进化和wiard这个方法进行清晰的定义
\subsection{Ablation Experiment}
Our method can be proved effective in two key areas: expanding the action space and using MCTS evolution. As shown in Table \ref{Ablation-study}, models with expanded action space (denoted as + methods) consistently outperform those without it, regardless of using random or MCTS evolution. For example, the Mistral using the MCTS+ method shows a 3.72\% improvement over the Evol-Instruct method. Additionally, data evolved through MCTS maintains high quality, further improving the instruction-following abilities of the model. The Phi-3 model, using MCTS evolution, improves performance by 1.5\% before action space expansion and by 2.92\% after expansion.

\section{Conclusion}
In this paper, we introduce a novel framework that leverages the power of MCTS combined with heuristic evaluation to synthesis high-value instruction data. Our statistical analysis validates the framework's effectiveness in synthesizing high-value data. By fine-tuning open-source models with these evolved instructions, models achieve competitive competitive performance compared to previous methods. 
\section*{Limitations}
We need to acknowledge that the process of using LLMs for evolving instructions with MCTS is opaque and incurs API costs. Knowledge distillation might balance the trade-off between expenses and synthesizing high-quality instructions. On the other hand, we have demonstrated the effectiveness of MCTS-evolved instructions under low-resource conditions. Further exploration of scaling laws could enhance our understanding of the framework.

\section*{Acknowledgments}

This work was supported by the Zhejiang Provincial Natural Science Foundation of China under Grant No. LZ23F020009, the NSFC project (No.~62072399), MoE Engineering Research Center of Digital Library, China Research Centre on Data and Knowledge for Engineering Sciences and Technology, the Fundamental Research Funds for the Central Universities (No.~226-2024-00170), and Ant Group. We also express our sincere gratitude to anonymous reviewers for their invaluable feedback and constructive comments.

\bibliography{anthology,custom}
\clearpage
\appendix
\section{Benchmark Details}
\label{Appendix_benchmark}
\begin{figure*}[ht]
    \centering
    \includegraphics[width=1.0\textwidth]{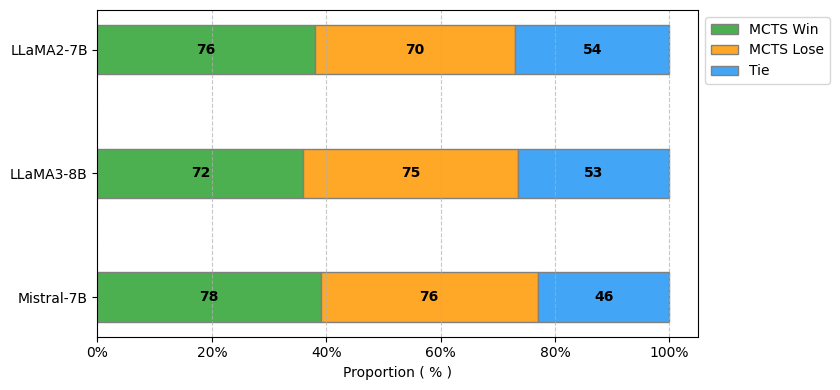}
    \caption{Manual evaluation of the results on Alpaca-eval.}
    \label{fig:human}
\end{figure*}
Alpaca-Eval \cite{alpaca_eval} is a comprehensive evaluation framework incorporating examples from diverse datasets, including self-instruct \cite{wang2022self}, open-assistant \cite{kopf2304openassistant}, Vicuna \cite{chiang2023vicuna} and Koala \cite{koala_blogpost_2023}. This framework uses English instructions across multiple categories and tasks to evaluate model performance in real-world scenarios.

MT-Bench \cite{zheng2024judging} is a benchmark designed to assess models' multi-turn conversational and instruction-following abilities. It contains 80 high-quality, multi-turn questions that represent common use cases. The development of MT-Bench is informed by eight categories of user prompts: writing, roleplay, extraction, reasoning, math, coding, stem knowledge, and humanities/social sciences knowledge.
\begin{figure*}
    \centering \includegraphics[width=1.0\linewidth]{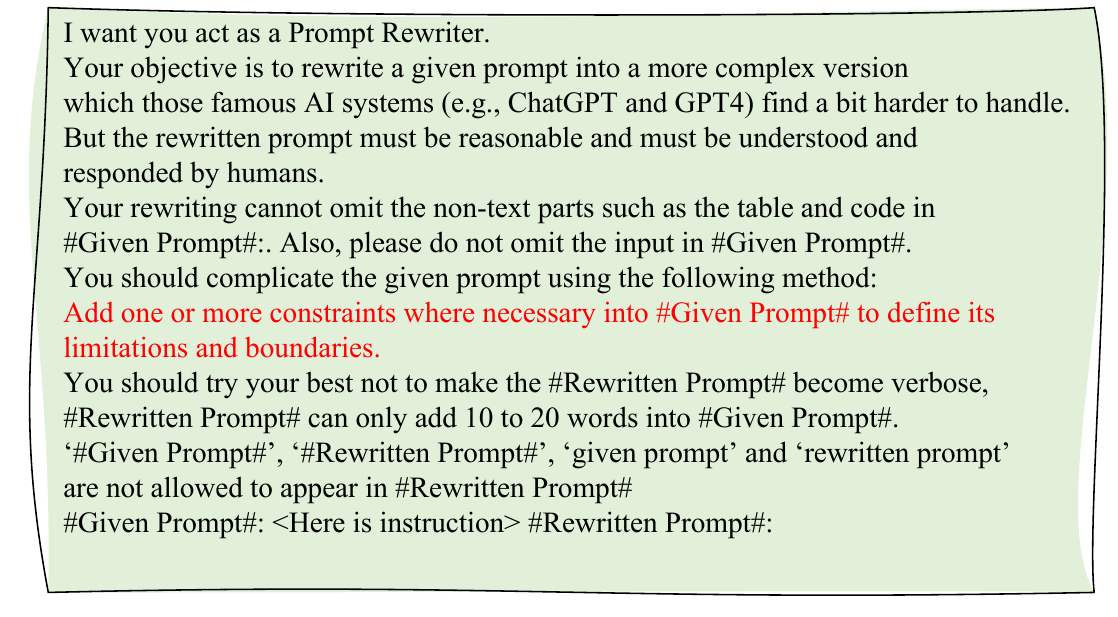}
    \caption{The Evolution Prompt: Add Key Constraints.}
    \label{fig:contraints}
\end{figure*}

\begin{table*}[ht]
    \centering
    \renewcommand{\arraystretch}{1.2} % Adjust row height
    \adjustbox{width=\textwidth}{
        \begin{tabular}{|>{\centering\arraybackslash}m{0.3\textwidth}|>{\arraybackslash}m{0.65\textwidth}|}
            \hline
            \textbf{Evolution Action} & \textbf{Instruction Evolution Case} \\ \hline
            
                  Add Global and Local Goals & 
        Original: "Please help me write an article about climate change." \newline
        Evolved: "Please help me write an article about climate change, aiming to educate readers on the causes and effects of climate change and suggest individual actions to combat it." \\ \hline
        
        Add Key Constraints & 
        Original: "Please help me design a website." \newline
        Evolved: "Please help me design a website that supports mobile access, and meets accessibility standards." \\ \hline
        
        Add Task Requirements & 
        Original: "Please help me prepare a meeting report." \newline
        Evolved: "Please help me prepare a meeting report including next quarter's sales strategy recommendations." \\ \hline
        
        Add Problem-Solving Skills & 
        Original: "Please help me solve this math problem. " \newline
        Evolved: "Please help me solve this math problem and explain each step and the mathematical principles used. " \\ \hline
        
        Add Reasoning Complexity & 
        Original: "Please provide some productivity tips." \newline
        Evolved: "Please provide some productivity tips, considering different work environments. Explain why these tips are effective." \\ \hline
        
        Add Domain Knowledge & 
        Original: "Please explain blockchain technology." \newline
        Evolved: "Please provide a detailed explanation of blockchain technology from an IT perspective, including its principles, applications, and future trends." \\ \hline
        
        Add Life Topics & 
        Original: "Please give me some healthy eating advice." \newline
        Evolved: "Please give me some healthy eating advice, especially for people who exercise regularly." \\ \hline
        
        Add Real-World Applications & 
        Original: "Please explain artificial intelligence." \newline
        Evolved: "Please explain artificial intelligence and provide examples of its applications in education, customer service, and business." \\ \hline
        
        Add Emotional Expression & 
        Original: "Please help me plan a trip." \newline
        Evolved: "Please help me plan an exciting trip, including some unique attractions and experiences to ensure the journey is fun." \\ \hline
        
        Format the Input Style & 
        Original: "Please give me some investment advice." \newline
        Evolved: "As a financial advisor, please give me some stock investment advice, especially beginner strategies." \\ \hline
        
        Format the Output Style & 
        Original: "Please summarize this article." \newline
        Evolved: "Please summarize this article in a report format, including main points, supporting data, and conclusions." \\ \hline
        
        Refine the Factuality & 
            Original: "Please describe the process of recycling." \newline
            Evolved: "Please accurately describe the process of recycling plastic bottles, including the collection, sorting, cleaning, shredding, and reprocessing steps." \\ \hline
        
        Create a New One & 
        Original: "Please give me some time management advice." \newline
        Evolved: "Please give me some advice on how to manage time efficiently throughout the day." \\ \hline
        
    \end{tabular}
}

\caption{Examples of Evolution Action}
\label{table:instruction_evolution}

\end{table*}

\section{Evolution Prompts}
\label{evol_prompt}
We designed the evolution prompts to serve as the action space. As shown in Figure \ref{fig:contraints}, it demonstrates a complete evolution prompt. By adding 10-20 words at each step, we ensure the iterative enhancement of the instruction data. 
Additionally, we presented the case of evolution action, as shown in \ref{table:instruction_evolution}.
\section{Human Eval}
\label{human_eval}

We conducted a blind pairwise comparison between two models: one trained on data generated by MCTS and the other on data generated through random evolution (Evol-Instruct). For this evaluation, we recruited 3 well-educated annotators. Each annotator was presented with two responses: one from the MCTS-based model and one from the random evolution-based model, with their sources randomly shuffled to ensure anonymity. The annotators evaluated each response based on the following criteria \cite{xu2023wizardlm}: (1) Relevance, (2) Knowledgeability, (3) Reasoning, (4) Calculation, and (5) Accuracy. They judged which response was superior for each comparable instance. To estimate the win rate, we compared the frequency of model wins with MCTS. As shown in Figure \ref{fig:human}, the model trained on MCTS-generated data achieved significantly better results than the model trained on randomly evolved data. This demonstrates the effectiveness of the MCTS method.

\begin{table*}[h!]
    \centering
      \adjustbox{width=0.9\textwidth}{
    \begin{tabular}{|>{\raggedright\arraybackslash}p{0.4\linewidth}|>{\raggedright\arraybackslash}p{0.5\linewidth}|}
        \hline
        \textbf{Seed Instruction} & \textbf{Evolved Instruction} \\ \hline
         \multirow{4}{=}{Create a list of ingredients to make a traditional lasagna.}  & Can you compile a tantalizing list of ingredients for both a traditional and a vegetarian lasagna recipe, each with no more than 10 ingredients, ensuring they are both delicious and visually appealing? \\ \hline
         
        \multirow{3}{=}{Generate a conversation between two friends talking about a dream they shared}  & Generate a detailed conversation between two friends talking about a vivid and exciting dream they shared last night. \\ 
         \hline

        \multirow{3}{=}{Create a list of ten shared characteristics between birds and reptiles.}  & Can you generate a list of ten common traits shared by avian and reptilian species, utilizing biological taxonomy and comparative anatomy? \\ 
         \hline
         
        \multirow{8}{=}{Name five famous French writers.}  & Imagine you are preparing a presentation for a literary seminar. Can you name five famous French writers from the 19th and 20th centuries, provide brief biographical information for each, and cite at least one notable work? Emphasize their contributions to French culture and literature, and highlight the impact of their works on the global literary community. \\ 
         \hline
        \multirow{6}{=}{What are the best methods for controlling finances?}  & Can you identify and implement the most effective and sustainable methods for managing personal finances in today's technology-driven society? This should include detailed budgeting techniques and saving strategies that leverage modern financial tools and apps. \\ 
         \hline
    \end{tabular} }
    \caption{Seed and Evolved Instructions with MCTS.}
    \label{table:instructions}
\end{table*}
\begin{figure*}
    \centering \includegraphics[width=0.9\linewidth]{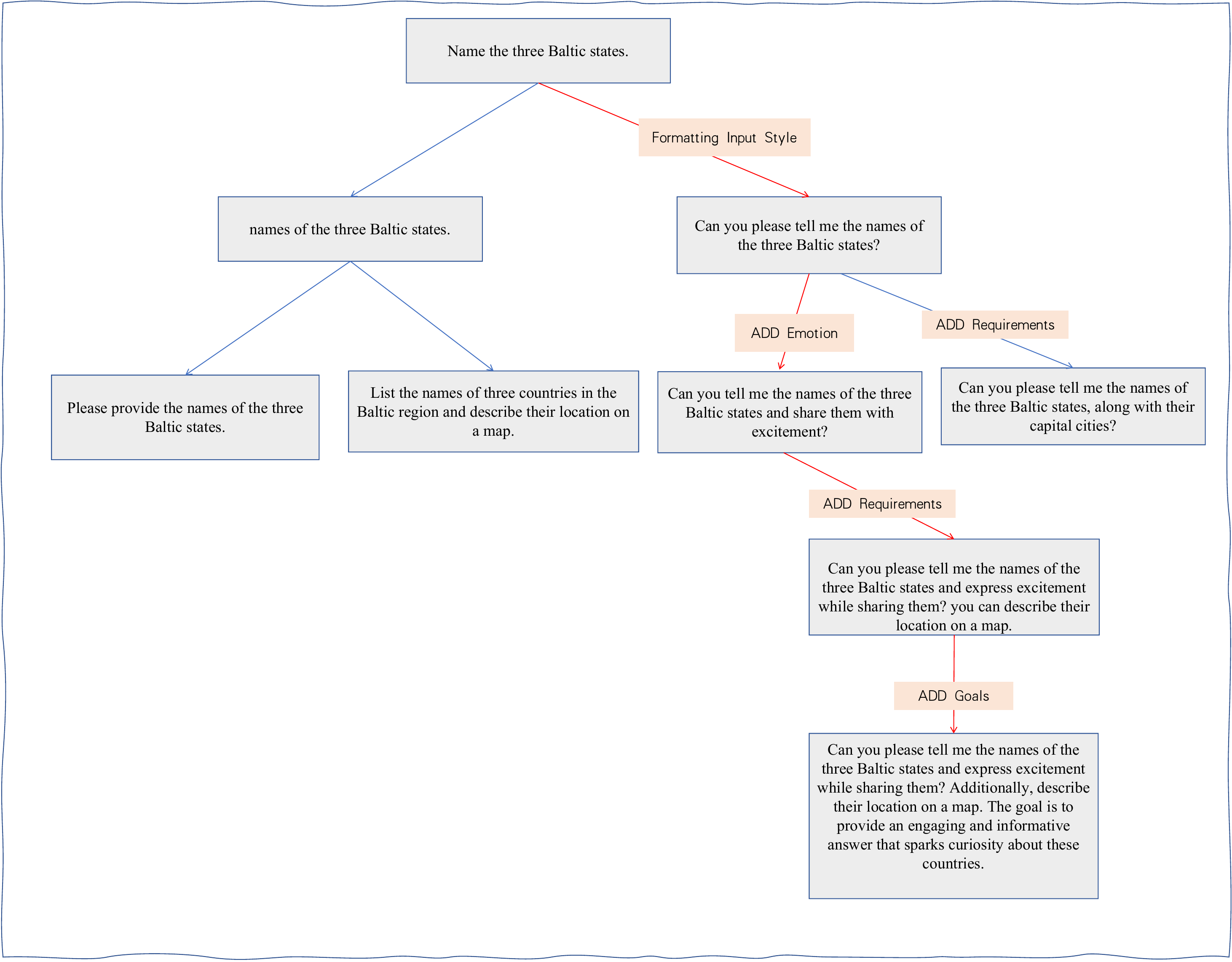}
    \caption{A Case of Instruction Evolution with MCTS.}
    \label{fig:case}
\end{figure*}
\section{Case Study}
\label{case_study}
We present a case study in Figure \ref{fig:case} to show the iterative evolution of a seed instruction. Starting with the seed instruction, "Name the three Baltic states," we progressively refine it to, "Can you please tell me the names of the three Baltic states and express excitement while sharing them? You can also describe their location on a map.". This process, guided by evaluation models, enhances the efficiency of evolving instructions. High-value instructions are identified and used as the basis for further evolution. Examples of instructions before and after the evolution are provided in Table \ref{table:instructions}.

\end{document}